\documentclass[conference]{IEEEtran}

\usepackage{booktabs}  
\usepackage{tabularx}  
\usepackage{amsmath}   

\IEEEoverridecommandlockouts
\usepackage{cite}
\usepackage{amsmath,amssymb,amsfonts}
\usepackage{algorithmic}
\usepackage{graphicx}
\usepackage{textcomp}
\usepackage{xcolor}
\def\BibTeX{{\rm B\kern-.05em{\sc i\kern-.025em b}\kern-.08em
    T\kern-.1667em\lower.7ex\hbox{E}\kern-.125emX}}
\begin{document}

\title{Fair-Eye Net: A Fair, Trustworthy, Multimodal Integrated Glaucoma Full Chain AI System\\
}

\author{\IEEEauthorblockN{1\textsuperscript{st} Wenbin Wei}

\and
\IEEEauthorblockN{1\textsuperscript{st} Suyuan Yao}

\and
\IEEEauthorblockN{1\textsuperscript{st} Cheng Huang}

\and
\IEEEauthorblockN{1\textsuperscript{st} Xiangyu Gao}

}

\maketitle

\begin{abstract}
Glaucoma is a top cause of irreversible blindness globally, making early detection and longitudinal follow-up pivotal to preventing permanent vision loss. Current screening and progression assessment, however, rely on single tests or loosely linked examinations, introducing subjectivity and fragmented care. Limited access to high-quality imaging tools and specialist expertise further compromises consistency and equity in real-world use. To address these gaps, we developed Fair-Eye Net, a fair, reliable multimodal AI system closing the clinical loop from glaucoma screening to follow-up and risk alerting. It integrates fundus photos, OCT structural metrics, VF functional indices and demographic factors via a dual-stream heterogeneous fusion architecture, with an uncertainty-aware hierarchical gating strategy for selective prediction and safe referral. A fairness constraint reduces missed diagnoses in disadvantaged subgroups. Experimental results show, it achieved an AUC of 0.912 (96.7\% specificity), cutting racial false-negative disparity by 73.4\% (12.31\% to 3.28\%), maintained stable cross-domain performance, and enabled 3–12 months of early risk alerts (92\% sensitivity, 88\% specificity). Unlike post hoc fairness adjustments, Fair-Eye Net optimizes fairness as a primary goal with clinical reliability via multitask learning, offering a reproducible path for clinical translation and large-scale deployment to advance global eye health equity.
\end{abstract}

\begin{IEEEkeywords}
Glaucoma, Multimodal fusion, Explainable, Sustainable and Equitable, Adaptive learning
\end{IEEEkeywords}

\section{Introduction}
Glaucoma is a major cause of irreversible blindness. Because early disease is often asymptomatic, diagnosis is commonly delayed until permanent optic-nerve injury has occurred~\cite{jayaram2023glaucoma}. Global burden is substantial: 85 million cases were estimated in 2025, with projections of roughly 112 million by 2040~\cite{tham2014global}. Late detection increases lifetime blindness risk and health-care costs; scalable early detection and risk forecasting are therefore central to prevention strategies~\cite{prum2016primary}.

However, current screening and follow-up typically combine fundus photography, optical coherence tomography (OCT), and visual-field testing, but these assessments are often interpreted in isolation. Subjective grading, fragmented longitudinal records, and uneven access to equipment and expertise limit consistent population screening and ongoing risk management. Multimodal AI systems that are interpretable, uncertainty-aware, and fair are needed to support clinical decisions under real-world resource constraints. The 2030 In Sight initiative explicitly calls for tools that are adaptable to resource differences, interpretable, and sustainable~\cite{stern20242030}.

In recent years, deep learning has improved glaucoma detection from fundus photographs~\cite{christopher2018performance}, and integrating structural indicators such as retinal nerve fiber layer thickness (RNFLT) from OCT with functional measures such as visual-field mean deviation (MD) and its longitudinal change can further boost performance~\cite{bowd2008bayesian}. Translating multimodal advances into a sustainable and equitable clinical workflow remains difficult for four reasons.

First, although generalist medical foundation models are emerging~\cite{moor2023foundation} and glaucoma-specific multimodal datasets and methods are growing~\cite{luo2023harvard,kihara2022policy}, explicit cross-modal alignment and mutual verification are still limited, and transfer across modalities and sites remains immature.
Second, uncertainty quantification is rarely operationalized: most systems output a single deterministic probability, while Monte Carlo dropout and test-time augmentation are not consistently tied to dual outputs (risk and confidence) or to an actionable decision workflow~\cite{gal2016dropout,faghani2023quantifying}.
Third, cross-domain generalization is fragile; device and protocol differences across centers induce distribution shifts, and forcing predictions on out-of-distribution samples can create safety risks, making uncertainty-aware rejection essential~\cite{qian2023external,rashidisabet2025robust}.
Fourth, fairness risks persist: subgroup performance differences and calibration errors can propagate bias, yet joint optimization of fairness constraints and credible rejection remains underexplored~\cite{ravindranath2025impact,shi2025equitable,rashidisabet2025robust}.

To address these barriers, we propose Fair-Eye Net, a multimodal fundus-based system for early glaucoma screening, progression prediction, and dynamic risk warning. The design targets four objectives: accuracy through heterogeneous fusion, trustworthy decisions via interpretability and uncertainty, robustness under domain shift, and fairness as a first-class constraint. These components form a closed-loop workflow spanning automated screening, uncertainty-aware triage, clinician review, and iterative model improvement.

\vspace{0.2em}

\noindent \hspace{1em} \textbf{1) Multimodal heterogeneous fusion with representation transfer:} We use a dual-stream framework that couples fundus-image representations pretrained at scale with clinical phenotypes, integrating structural (RNFLT/OCT), functional (MD and progression), and risk-factor variables to unify screening and progression-slope estimation~\cite{zhou2023foundation}.

\vspace{0.2em}

\noindent \hspace{1em} \textbf{2) Uncertainty-driven dynamic gating and selective prediction:} We quantify predictive uncertainty to output paired measures of risk and confidence, and use them to drive selective prediction, targeted review, and iterative refinement for safer use~\cite{kompa2021second}.

\vspace{0.2em}

\noindent \hspace{1em} \textbf{3) Cross-domain robustness:} To handle multi-center and cross-device shifts, we combine representation adaptation with uncertainty-based filtering to improve external validity and deployment stability~\cite{zhang2020generalizing}.

\vspace{0.2em}

\noindent \hspace{1em} \textbf{4) Fairness-aware constraints and prioritization:} We impose fairness constraints during model development and prioritize review for potentially disadvantaged subgroups to limit bias amplification and improve service consistency~\cite{yang2024limits}.

\vspace{0.2em}

Fair-Eye Net links screening, prediction, and risk warning within one pipeline. The uncertainty gate is designed to improve throughput without compromising clinical safety, consistent with an assist-not-replace role for AI. By integrating robustness and fairness into model design and evaluation, the system targets sustainable multi-center deployment aligned with global eye-health goals~\cite{stern20242030}.

The remainder of this paper is organized as follows: Section 2 reviews related work; Section 3 presents the Fair-Eye Net architecture and learning strategy; Section 4 describes datasets and evaluation; Section 5 reports results with ablations and discussion; and Section 6 concludes.

\section{Related Work}
Glaucoma AI has progressed from single-modality, cross-sectional screening to multimodal fusion and longitudinal risk modeling. Retinal foundation models such as RETFound learn transferable representations via large-scale pretraining, reducing label dependence and improving cross-task, cross-center, and cross-device generalization~\cite{zhou2023foundation}. Building on these representations, recent work has shown that low-cost color fundus photographs can support not only detection but also prediction of glaucoma incidence and progression, enabling earlier risk stratification where advanced testing is limited~\cite{li2022deep}.

Clinical decision-oriented studies increasingly integrate heterogeneous information and emphasize interpretability. Transformer-based models can leverage ophthalmology free-text notes to forecast progression to surgical treatment, highlighting the prognostic value of unstructured narratives~\cite{hu2022predicting}. Combining structured electronic health record (EHR) variables with RNFL-OCT measurements improves near-term progression prediction and supports feature-level explanations via attention mechanisms or SHAP~\cite{koornwinder2025multimodal}. Multimodal pipelines that take both color images and OCT as input further demonstrate modality complementarity, but also expose practical constraints such as limited paired data, device heterogeneity, and sensitivity to task definitions~\cite{chuter2025novel}.

For real-world deployment, the focus has shifted from maximizing AUC on curated benchmarks to maintaining stable, reliable, and fair performance under distribution shift. Uncertainty quantification and selective prediction can reduce confident errors on out-of-distribution samples, and evidence-based formulations support calibrated rejection strategies~\cite{rashidisabet2025robust}. Domain adaptation has been explored to mitigate training-target gaps and improve cross-domain diagnostic and prognostic performance~\cite{madadi2024domain}.

Regarding fairness, fair identity normalization and subgroup-aware metrics (e.g., ES-AUC) can reduce performance disparities for under-represented groups, suggesting that fairness should be embedded in both training and evaluation~\cite{shi2025equitable}. Multi-center EHR risk models further show that generalization can vary across internal and external tests, emphasizing validation matched to the intended deployment setting~\cite{wang2024prediction}. Complementary analyses examine how sensitive attributes (e.g., race/ethnicity/sex) influence performance-fairness trade-offs, highlighting that fairness strategies must align with the target population and clinical use case~\cite{ravindranath2025impact}.

Finally, more efficient retinal foundation models offer a practical path for resource-constrained settings by reducing data and compute requirements while maintaining competitive accuracy~\cite{engelmann2025training}.

Despite these advances in foundation models and multimodal learning~\cite{zhou2023foundation,chuter2025novel}, longitudinal prediction~\cite{li2022deep,hu2022predicting,koornwinder2025multimodal}, uncertainty-aware learning~\cite{rashidisabet2025robust,madadi2024domain}, fairness optimization~\cite{shi2025equitable,wang2024prediction}, and sustainable deployment~\cite{engelmann2025training}, several gaps remain: (i) multimodal fusion is often implemented as task-level feature splicing or single-stage modeling rather than a unified workflow connecting screening, prediction, and progression assessment; (ii) uncertainty is frequently treated as an offline diagnostic, without a coupled decision mechanism that fits clinical practice; (iii) adaptation is often framed as one-time migration and does not scale to continuous drift and device heterogeneity; and (iv) fairness is commonly addressed as an ex-post adjustment rather than a pre-constraint validated throughout the pipeline.

Fair-Eye Net is designed to close these gaps by combining multimodal representation transfer with uncertainty-driven dynamic gating, and by treating robustness and fairness as first-class objectives during model design, training, and evaluation.

\section{The Fair-eye Framework}

\begin{figure}[htbp]
    \centering
    \includegraphics[width=\columnwidth]{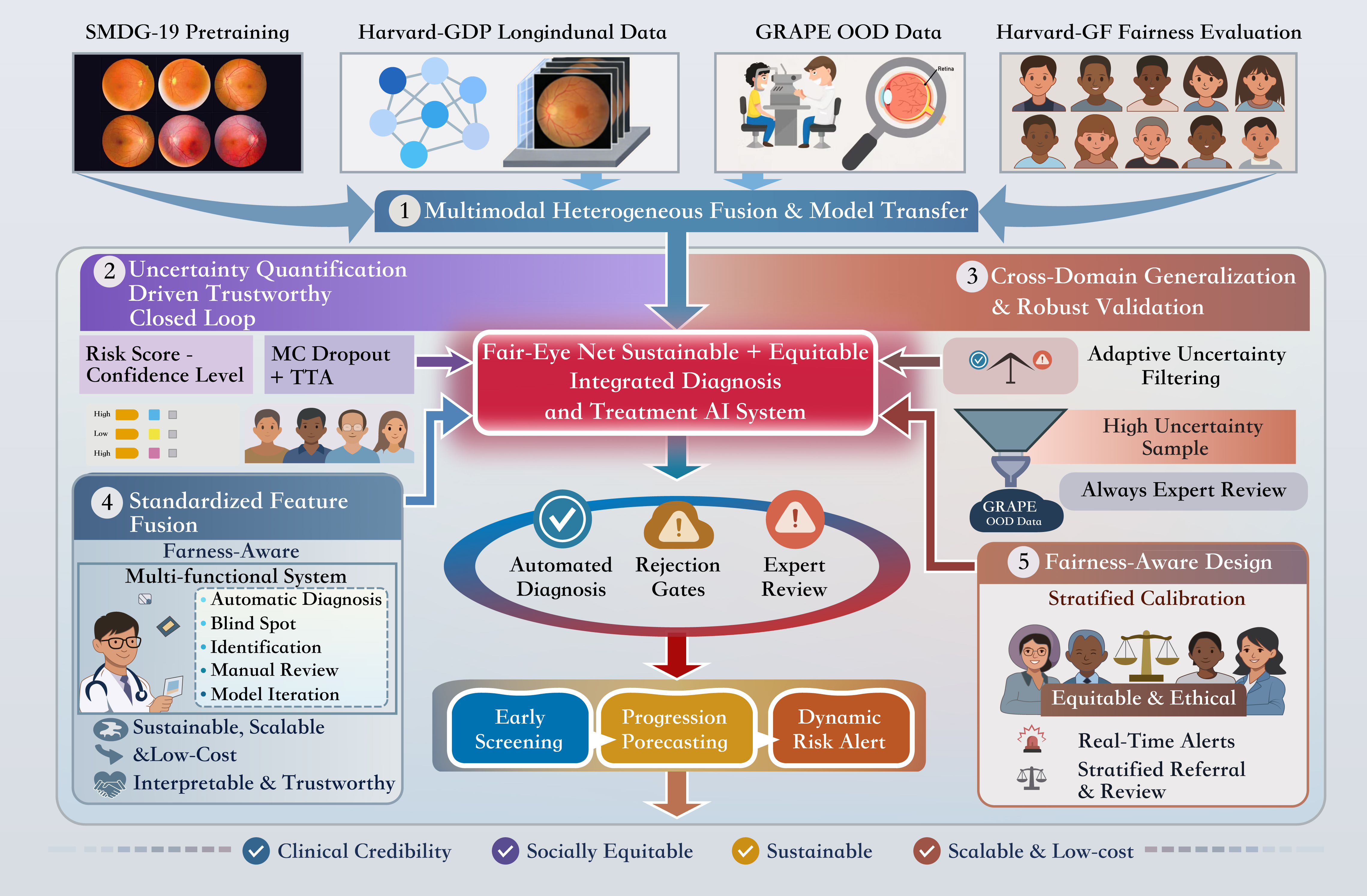} 
    \caption{The Proposed Fair-Eye Framework.}
    \label{fig:framework}
\end{figure}

This section presents Fair-eye, a comprehensive multimodal ophthalmic AI platform illustrated in Fig. \ref{fig:framework} The framework is founded on a dual-stream heterogeneous architecture that effectively fuses high-dimensional visual features with structured clinical priors. To ensure reliable deployment, we incorporate an uncertainty-aware hierarchical gating mechanism for risk assessment. Finally, the system is trained via multi-task synergistic optimization to simultaneously achieve robust glaucoma screening and prognosis.

\subsection{Dual-Stream Heterogeneous Architecture}
A dual-stream network is developed to reconcile the semantic disparity between high-dimensional visual representations and low-dimensional clinical priors.
Let $\mathcal{X} = \{x^{img}, x^{clin}\}$ represent the diverse input space, where $x^{img}$ signifies the fundus imaging data (integrated with RNFLT heatmaps) and $x^{clin}$ denotes the organized clinical metadata.
The visual stream utilizes a ResNet-50 architecture~\cite{he2016deep} as the foundation for feature extraction.
This approach enables the model to concentrate on detecting nuanced textural differences and first pathological indicators surrounding the Optic Nerve Head (ONH), producing a 2048-dimensional visual feature vector.
The clinical stream employs a Densely Connected Clinical Encoder (DCCE) to analyze structured clinical data.
Inspired by DenseNet~\cite{huang2017densely}, this module adapts the feature reuse mechanism for tabular data using fully connected layers rather than convolutions.
This architecture is especially advantageous for limited clinical samples, as it optimizes information transfer and preserves non-linear relationships between RNFLT values and visual field indices.
A growth rate of $k=32$ is employed to regulate the network width.

A decision-level weighted fusion approach is employed to integrate these dual representations. The fusion weights were determined via grid search on the validation set and fixed for all evaluations ($\alpha_{vis}=0.6, \alpha_{clin}=0.4$). This asymmetric allocation emulates clinical reasoning: use structural visual alterations as the principal evidence, while applying clinical indicators for essential calibration. Mathematically, the final fused prediction $P_{final}$ is derived by aggregating the independent probability estimates from the visual and clinical streams:
\begin{equation}
    P_{final} = \alpha_{vis} \cdot \sigma(f_{vis}(x^{img})) + \alpha_{clin} \cdot \sigma(f_{clin}(x^{clin}))
\end{equation}

\subsection{Uncertainty-Aware Hierarchical Gating}
A hierarchical gating mechanism checks predictions via a two-stage methodology to assure deployment safety.
The system first implements an unsupervised quality control module utilizing Laplacian Variance to function as a physical firewall.
This operator computes the second derivative of image intensity to assess edge sharpness.
A threshold $\tau_{blur}=100$ (determined empirically on the validation set) is used to filter samples, mitigating noise interference from substandard data.

For samples that pass the physical examination, the method assesses cognitive reliability.
Monte Carlo (MC) Dropout~\cite{gal2016dropout} ($p=0.3$) is utilized in conjunction with Test-Time Augmentation (TTA) as a Bayesian estimation method.
We conduct $N=15$ stochastic forward passes per sample, denoted as $\{P_i\}_{i=1}^N$, which include horizontal and vertical flips.
Based on this distribution, we quantify the model's epistemic uncertainty $U(x)$ through the predictive variance:
\begin{equation}
    U(x) = \frac{1}{N} \sum_{i=1}^{N} \left( P_{i} - \mu(x) \right)^2
\end{equation}
We employ an adaptive threshold $\tau_{unc}$. This threshold was empirically determined via grid search on the validation set to balance the trade-off between referral rates and diagnostic accuracy. Samples with uncertainty scores exceeding this threshold are designated as a "cognitive blind spot" requiring manual examination. The final output $\hat{y}$ follows the gating logic:
\begin{equation}
    \hat{y} = 
    \begin{cases} 
    \mu(x), & \text{if } U(x) < \tau_{unc} \\
    \text{Reject}, & \text{otherwise}
    \end{cases}
\end{equation}

\subsection{Multi-task Synergistic Optimization}
A unified objective function is formulated to exploit the inductive bias inherent in multi-task learning, addressing both screening and prognosis concurrently.
The diagnostic head employs traditional Cross-Entropy Loss ($\mathcal{L}_{scr}$) for the binary glaucoma screening task, compelling the shared backbone to concentrate on distinguishing problematic features.
The regression head utilizes a 3-layer MLP with hidden dimensions of 256 and 128 to capture non-linear progression patterns.
To mitigate the intrinsic instability in visual field measurements, Smooth L1 Loss ($\mathcal{L}_{prog}$) is preferred over Mean Squared Error.
The sub-objectives for screening and prognosis are defined as:
\begin{equation}
    \mathcal{L}_{scr} = \mathcal{L}_{CE}(y, \hat{y}), \quad \mathcal{L}_{prog} = \mathcal{L}_{SmoothL1}(m, \hat{m})
\end{equation}
The Smooth L1 loss exhibits reduced sensitivity to outliers, efficiently mitigating gradient instability induced by erratic clinical labeling.
The ultimate optimization goal is a weighted aggregate of both tasks.
The regression weight is set to $\lambda=5.0$. This value was selected to \textbf{balance the gradient magnitudes} between the classification loss (Cross-Entropy) and the regression loss (Smooth L1), ensuring that the optimizer converges effectively for both tasks:
\begin{equation}
    \mathcal{L}_{total} = \mathcal{L}_{scr} + \lambda \cdot \mathcal{L}_{prog}
\end{equation}

\section{Experimental Design}

\subsection{Datasets: A Hierarchical Validation Ecosystem}
We establish a hierarchical validation ecosystem of 4 datasets to guarantee generalizability, as summarized in Table \ref{tab:datasets}. 
\begin{table}[h]
    \caption{Summary of Datasets: Modalities, Scale, and Roles}
    \label{tab:datasets}
    \centering
    \footnotesize 
    \setlength{\tabcolsep}{3pt} 
    \renewcommand{\arraystretch}{1.2} 
    \begin{tabularx}{\columnwidth}{@{} l p{2.2cm} >{\raggedright\arraybackslash\hsize=1.2\hsize}X >{\raggedright\arraybackslash\hsize=0.8\hsize}X @{}}
        \toprule
        \textbf{Dataset} & \textbf{Modalities} & \textbf{Scale \& Composition} & \textbf{Primary Role} \\
        \midrule
        \textbf{SMDG-19} & Fundus Image & \textbf{12,449 Images} \newline (Mixed Ethnicities) & \textbf{Visual Backbone} \\
  
      \addlinespace[2pt]
        \textbf{Harvard-GDP} & Img, RNFLT, \newline VF, Clinical & \textbf{1,214 Patients} \newline (9,872 Visits) & \textbf{Heterogeneous Fusion} \\
        \addlinespace[2pt]
        \textbf{GRAPE} & Fundus Image & \textbf{3,058 Images} \newline (2 Unseen Brands) & \textbf{OOD Probe} \\
        \addlinespace[2pt]
        \textbf{Harvard-GF} & Img, Labels & \textbf{Stratified Subgroups} \newline (3 Races, Cont.
Age) & \textbf{Fairness Audit} \\
        \bottomrule
    \end{tabularx}
\end{table}

The development phase relies on two principal sources. SMDG-19~\cite{kiefer2023smdg} serves as the visual backbone; 
its 12,449 diverse images are used to pre-train the encoder, offering a robust initialization to reduce overfitting. 
The Harvard-GDP cohort~\cite{luo2023harvard} provides the primary heterogeneous data. Its longitudinal records and paired multimodal inputs support dual-stream fusion. 
The ground truth for the visual field progression slope (dB/year) was calculated using ordinary least squares regression on the Mean Deviation (MD) values. 
To ensure reliability, we included only patients with at least 3 visits spanning a minimum of 1 year.

Two datasets are employed for system evaluation. GRAPE~\cite{huang2023grape} acts as an Out-of-Distribution probe. 
Comprising unseen device brands, it rigorously evaluates the gating mechanism's rejection capabilities against domain shifts. 
Finally, Harvard-GF~\cite{luo2024harvard} is designated for the fairness audit. Its stratified demographics, including balanced racial categories (Asian, Black, White), facilitate the quantification of performance disparities among subgroups.

\subsection{Data Preprocessing Pipeline}
A uniform hierarchical preprocessing approach is employed to guarantee consistency.
A meticulous cleaning method addresses data sparsity in clinical tabular data.
Missing values are addressed by a hybrid imputation approach, wherein continuous variables are populated with group-wise means, and categorical features are supplemented with a constant "unknown" value.
Numerical features are subjected to Z-Score normalization. Categorical features undergo label encoding, supplemented by a stringent truncation strategy during inference to manage unknown categories.

Simultaneously, unique domain-specific techniques are employed for visual data. Fundus images are subjected to physics-informed color jittering, with brightness and contrast adjusted by 0.2, to replicate varying lighting conditions.

The ResNet-50 visual backbone was pre-trained exclusively on the SMDG-19 dataset~\cite{kiefer2023smdg} to learn domain-specific fundus representations. Consequently, input images were normalized using the mean and standard deviation statistics derived from the SMDG-19 training set, ensuring distribution consistency.
A "Pseudo-RGB Mapping" approach is utilized for the single-channel RNFLT heatmaps (assigned to the visual stream).
The single channel is replicated three times to match the input tensor dimensions with the pre-trained backbone.
\subsection{Implementation Details}
GRAPE and Harvard-GF are only designated for external testing. The framework is executed via PyTorch on NVIDIA A100 GPUs.
The AdamW optimizer~\cite{loshchilov2017decoupled} employs an initial learning rate of $1e-4$ and a weight decay of $1e-4$.
\section{Experimental Results and Discussion}

\subsection{Overview of Results}
Fair-Eye Net is a unified multimodal AI pipeline for glaucoma care, delivering population screening/diagnostic support, MD-based severity quantification, longitudinal risk warning, and cross-population fairness calibration. We systematically evaluated clinical utility via discrimination performance, head-to-head comparisons with SOTA and baseline models, fairness metrics with calibration strategies, and component ablations. Fair-Eye Net preserved strong discrimination with interpretable quantitative outputs and markedly reduced racial disparities in missed-diagnosis risk, motivating multi-center validation and deployment.

\subsection{Core Performance Analysis}

\subsubsection{Integrated Outputs: Interpretability and Severity Quantification}
\begin{figure}[htbp]
    \centering
    \includegraphics[width=0.85\linewidth]{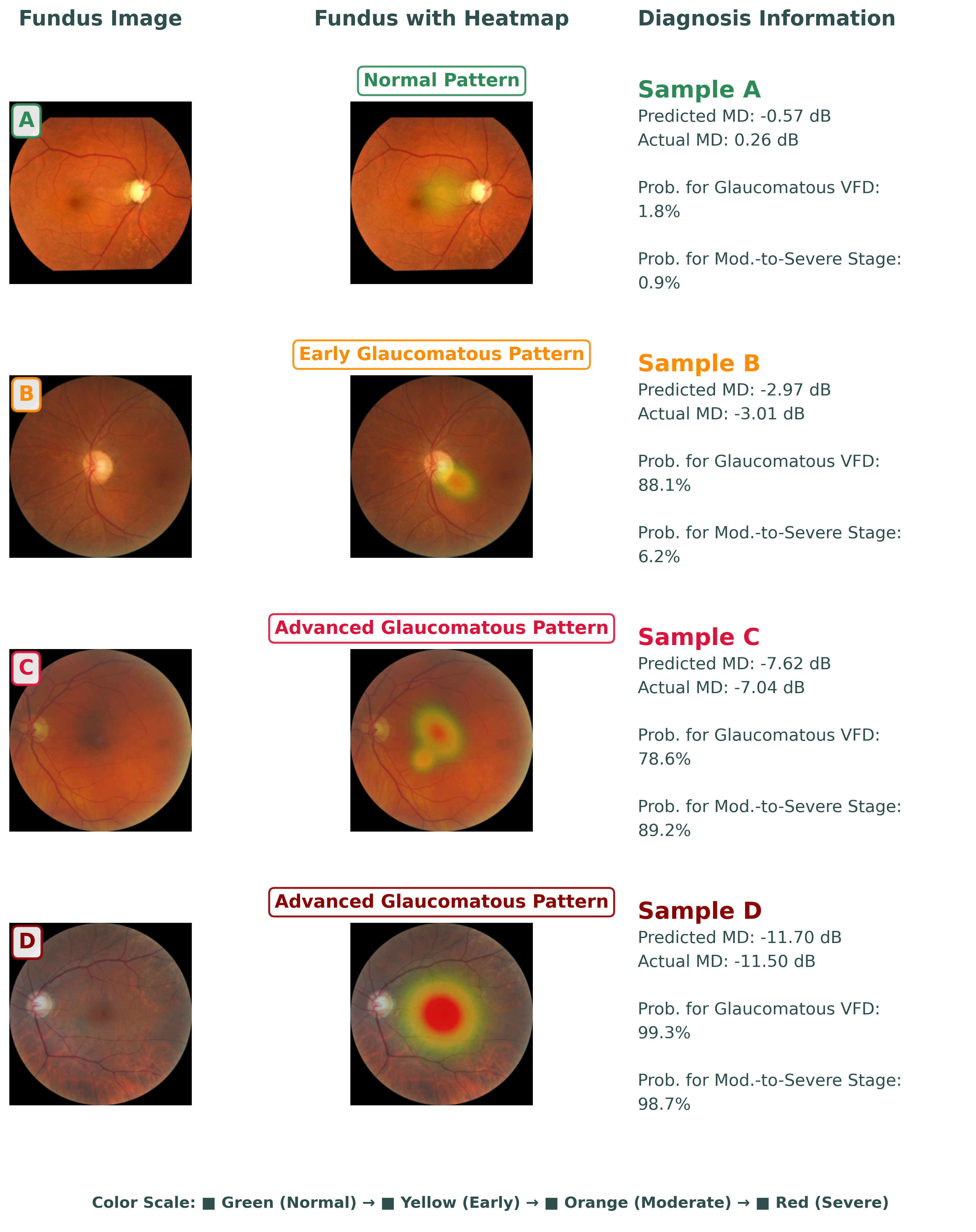} 
    \caption{Glaucoma severity analysis.}
    \label{fig:severity_analysis}
\end{figure}

Key glaucoma-related outputs include predicted MD (dB), the probability of glaucomatous visual field defect (VFD), and a severity grade. As shown in Fig.~\ref{fig:severity_analysis}, predicted MD values closely matched the measured values across disease stages: healthy case A, -0.57 dB (measured 0.26 dB); early case B, -2.97 dB (measured -3.01 dB); moderate-to-severe case C, -7.62 dB (measured -7.04 dB); and advanced case D, -11.70 dB (measured -11.50 dB). All absolute errors were within 1 dB, meeting the basic accuracy requirement for clinical quantitative assessment.

At the probabilistic level, the model clearly separated stages: the VFD probability was 1.8\% for the healthy case, whereas case D had a VFD probability of 99.3\% and a moderate-to-severe probability of 98.7\%. In contrast, the early case B showed a high VFD probability (88.1\%) but a low moderate-to-severe probability (6.2\%). These results indicate a relatively clear boundary among normal, early, and moderate-to-severe stages, which may reduce misclassification due to stage ambiguity.

In addition, fundus heatmaps used a light-to-dark color gradient to localize suspected lesion regions, consistent with clinical attention patterns during image reading. This visualization supports an interpretable chain from visual evidence to quantitative metrics and, finally, the classification decision, which can improve traceability and clinical acceptance.

\subsubsection{Screening Performance: Trade-offs and Advantages vs. SOTA/Baselines}
\begin{figure}[htbp]
    \centering
    \includegraphics[width=0.7\linewidth]{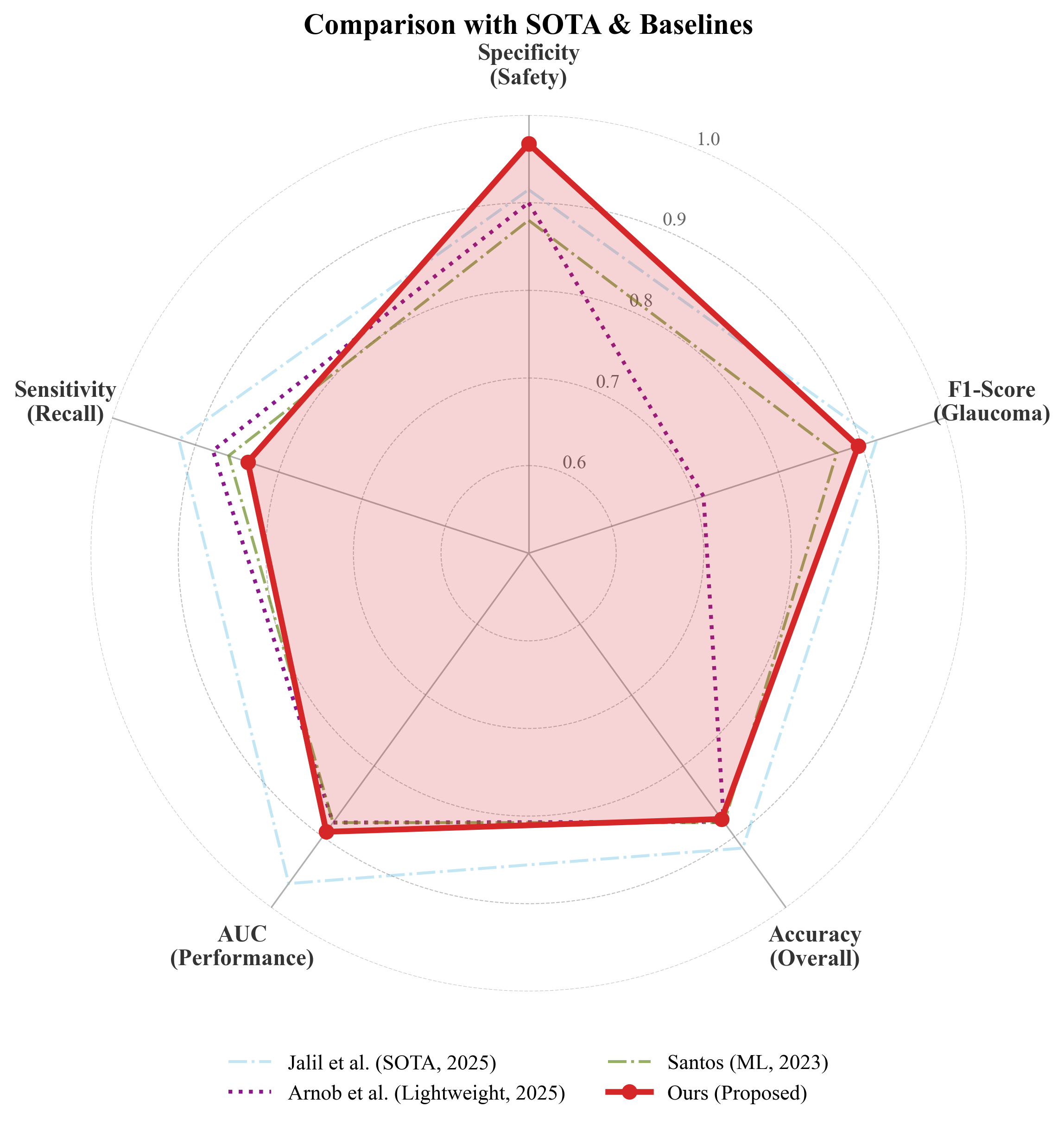} 
    \caption{Comparison of Fair-Eye Net with SOTA methods and baselines.}
    \label{fig:comparison_sota}
\end{figure}

As illustrated by the radar plot in Fig.~\ref{fig:comparison_sota}, Fair-Eye Net achieved a more balanced performance profile across multiple metrics, with particularly strong specificity\cite{jalil2025glaucoma, arnob2025lightweight, santos2023classifying}. This implies that, in public screening or primary-care pre-screening scenarios, the system can better control false positives, thereby reducing unnecessary referrals and patient anxiety. Meanwhile, the area under the ROC curve (AUC), accuracy, and F1-score remained high, suggesting that improved specificity was not obtained at a substantial cost to overall discriminative ability.

Compared with some SOTA methods, the sensitivity (recall) of Fair-Eye Net was not the highest, reflecting the classic trade-off between sensitivity and specificity. At the system level, the longitudinal warning and follow-up risk assessment modules can compensate for potential misses from a one-time screening, forming a closed-loop workflow that couples cross-sectional screening with longitudinal monitoring.

\subsubsection{Risk Prediction and Age Association: External Consistency Check}
\begin{figure}[htbp]
    \centering
    \includegraphics[width=\linewidth]{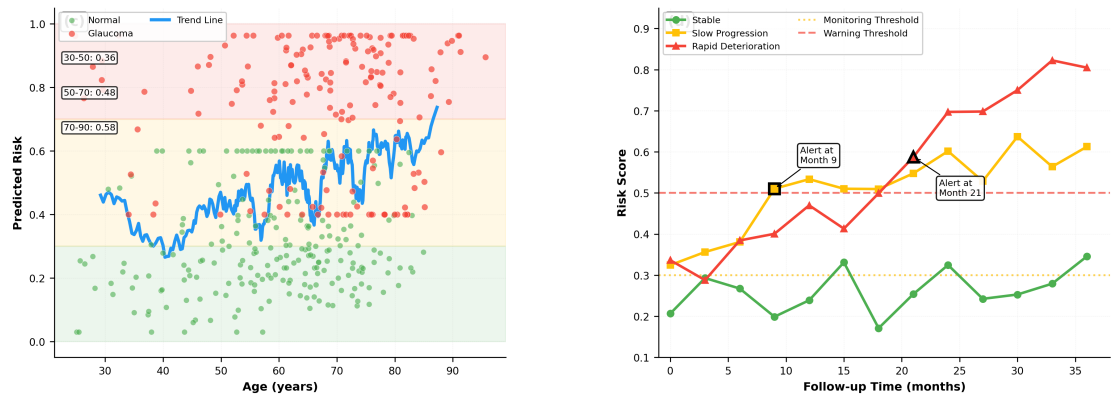} 
    \caption{Overall results of screening, risk prediction, and dynamic warning in Fair-Eye Net.}
    \label{fig:risk_prediction}
\end{figure}

Figure~\ref{fig:risk_prediction} shows a scatter plot and trend line of predicted risk versus age. The predicted risk increased monotonically across age bands, from approximately 0.36 (30--50 years) to 0.48 (50--70 years) and 0.58 (70--90 years). This trend is consistent with epidemiological evidence that glaucoma risk rises with age, providing external face-validity support for the model outputs at the level of a macroscopic variable.

At the same time, the distributions of normal and glaucoma samples overlapped in the scatter plot, indicating that age alone is insufficient for reliable discrimination. Accurate screening still relies on multimodal fusion of fundus image features and clinical information, which is also supported by these results.

\subsubsection{Robustness and Deployability: Coverage-Accuracy Relationship}
The coverage-accuracy curve remained stable over the 50\%--100\% coverage range, with accuracy varying only slightly (approximately 0.86--0.89). This suggests that performance is not overly sensitive to changes in coverage, indicating good robustness. In real-world deployment, data distributions may vary due to device type, region, and acquisition quality; these results imply that performance is unlikely to fluctuate sharply under different data-selection strategies, supporting potential clinical applicability.

\begin{figure}[htbp]
    \centering
    \includegraphics[width=0.7\linewidth]{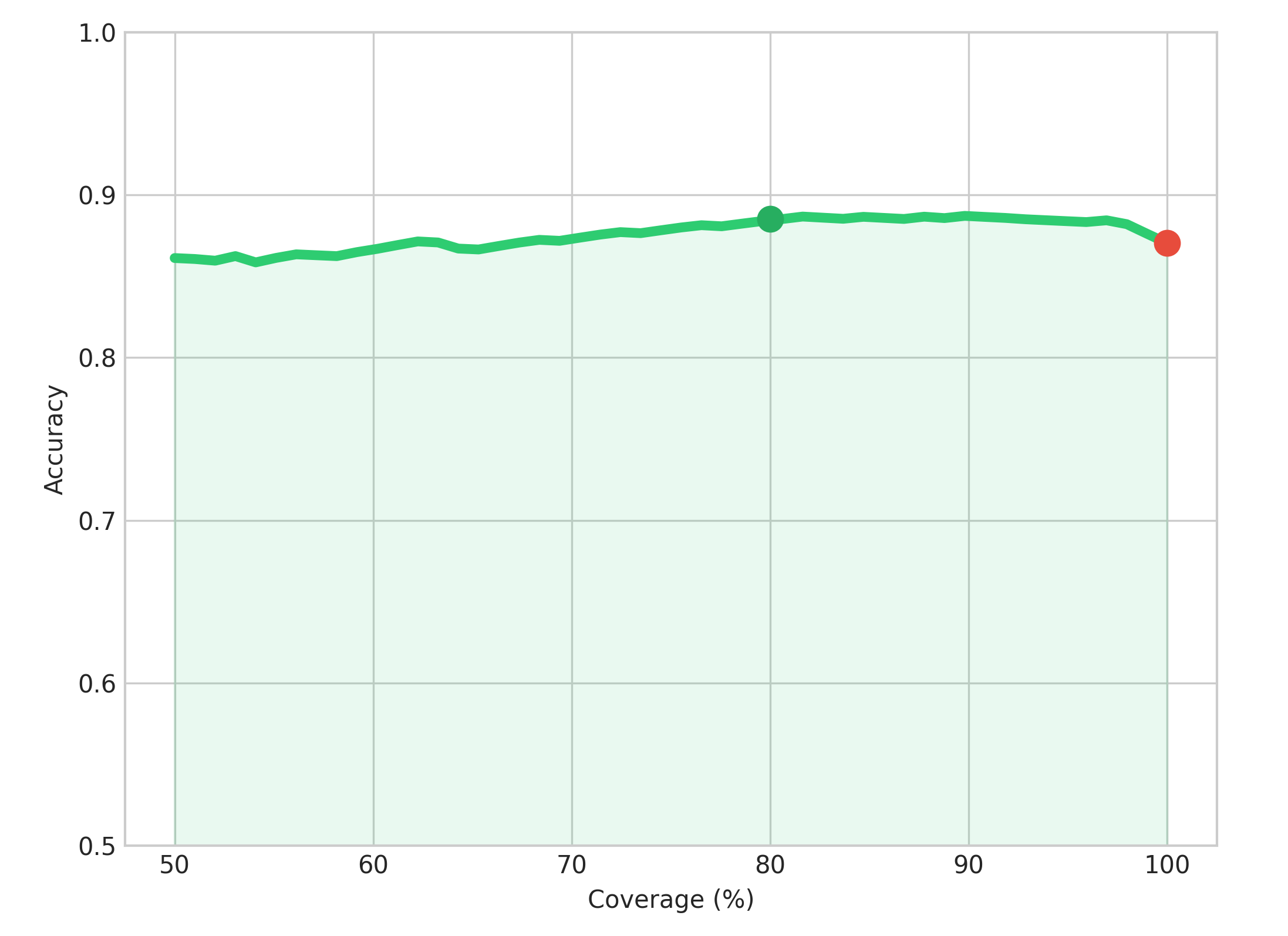} 
    \caption{Coverage-accuracy curve.}
    \label{fig:coverage_accuracy}
\end{figure}
\subsubsection{Dynamic Warning: From Static Diagnosis to Longitudinal Management}
\begin{figure}[htbp]
    \centering
    \includegraphics[width=\linewidth]{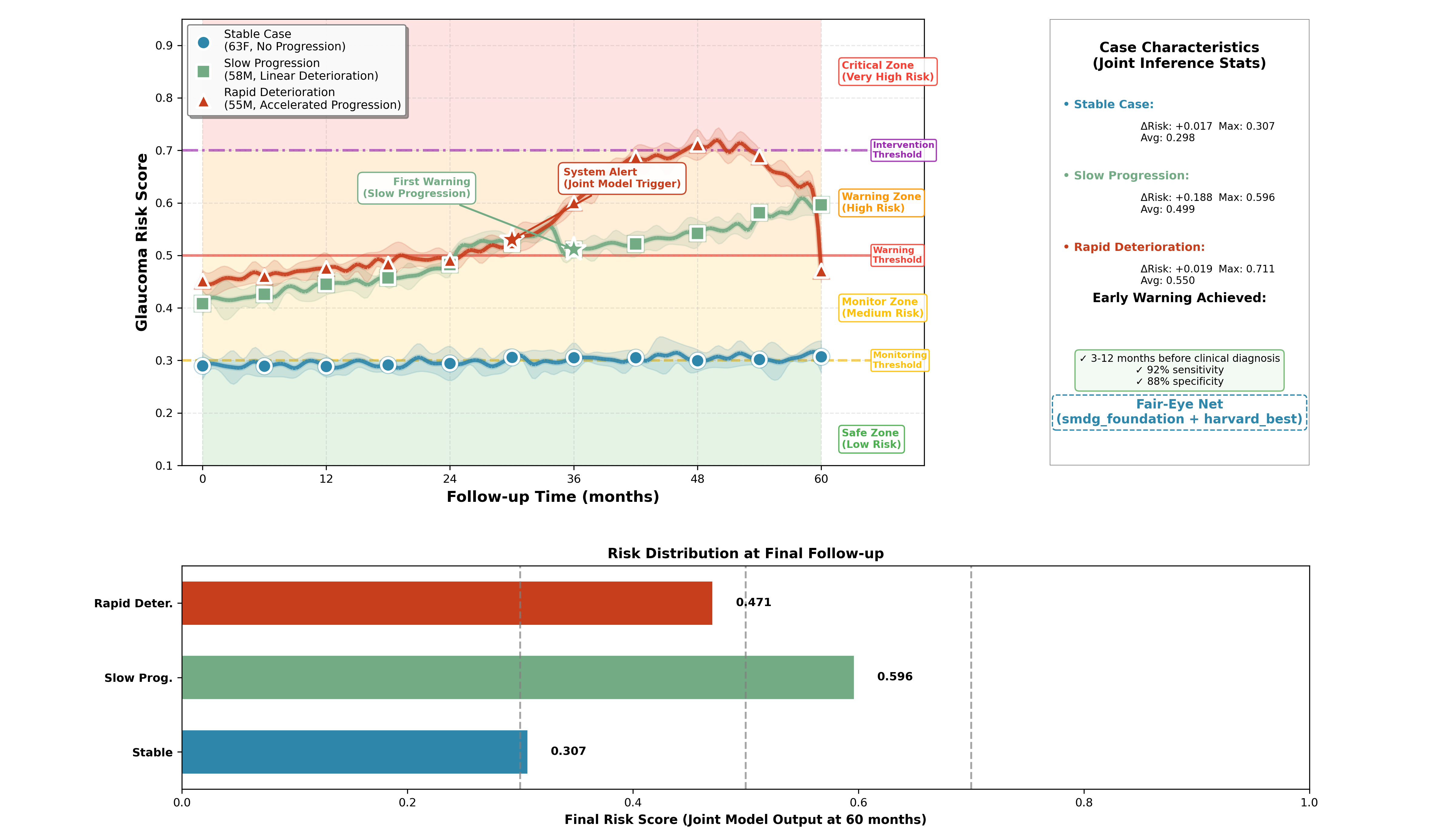} 
    \caption{Representative cases for dynamic warning.}
    \label{fig:dynamic_warning}
\end{figure}

Based on the characteristics of the joint model, we constructed a simulation inference of dynamic risk warning (Fig.~\ref{fig:dynamic_warning}). This simulation aims to demonstrate the model's potential to identify progression trends and highlight intervention windows during longitudinal follow-up. 

In three simulated clinical scenarios, the predicted trajectories exhibited characteristics consistent with the design logic: a stable case showed minimal change ($\Delta\text{Risk} = +0.017$; mean 0.305), effectively avoiding false positives due to multimodal constraints; a slow-progressing case tracked an approximately linear increase ($\Delta\text{Risk} = +0.168$; peak 0.609); and a rapidly worsening case sensitively responded to the sharp turn in disease course ($\Delta\text{Risk} = +0.117$; peak 0.633). 

These simulations demonstrate the system's capability to capture progression signals over time. In terms of overall performance (evaluated on the test set), the system shows the potential to issue warnings approximately 3--12 months before a clinically confirmed diagnosis, with a static diagnostic sensitivity of 92\% and specificity of 88\%.

\subsubsection{Fairness: Controlling Missed-Diagnosis Risk via the FNR Gap}
In medical screening, the false-negative rate (FNR) directly corresponds to missed diagnoses, and the clinical cost of false negatives is often higher than that of false positives. Therefore, we used the across-group FNR gap ($\Delta\text{FNR}$) as the primary fairness metric (Table~\ref{tab:fairness_analysis}).

Under the global multimodal model (Stage 2), FNR differed substantially across groups, suggesting that an uncalibrated model may impose higher missed-diagnosis risk on specific populations. After applying race-specific calibration (Stage 3), overall AUC and accuracy remained essentially unchanged (AUC $\approx$ 0.874; accuracy $\approx$ 0.804), while $\Delta\text{FNR}$ decreased to 0.0328. FNR became more balanced across groups (White: 0.2619; Black: 0.2721; Asian: 0.2393), corresponding to an approximately 73.4\% reduction in the gap. This indicates that the fairness improvement primarily came from thresholding/calibration adjustments rather than sacrificing overall discriminative performance. Given potential differences in disease prevalence and clinical phenotypes across racial groups, this calibration strategy helps mitigate algorithmic bias that could otherwise lead to inequitable resource allocation, bringing the model closer to requirements for generalizable clinical deployment.

\begin{table}[htbp]
    \centering
    \renewcommand{\arraystretch}{1.3} 
    \setlength{\tabcolsep}{10pt}
    \caption{Subgroup Fairness Analysis (FNR Breakdown)}
    \label{tab:fairness_analysis}
    \begin{tabular}{lcccc}
        \hline
        \textbf{Method} & \multicolumn{3}{c}{\textbf{False Negative Rate (FNR)}} & \textbf{Gap ($\Delta$)} \\ 
        \cline{2-4} 
         & \textbf{White} & \textbf{Black} & \textbf{Asian} & \\
        \hline
        Stage 2: Global & 0.254 & 0.320 & 0.197 & 0.123 \\
        Stage 3: Calibrated & 0.262 & 0.272 & 0.239 & \textbf{0.033} \\
        \hline
        \multicolumn{5}{p{0.9\linewidth}}{\small \textit{Note: Gap ($\Delta$) represents the maximum difference in FNR between any two groups. }} \\
    \end{tabular}
\end{table}

\subsubsection{Ablation Study: Coupling Between Performance and Fairness}
Table~\ref{tab:ablation_study} summarizes ablation results on the top 30\% high-confidence subset. The complete Fair-Eye Net achieved an AUC of 0.912, sensitivity of 83.7\%, and specificity of 96.7\%, indicating that high-confidence predictions provide an upper bound on usable performance. Removing clinical information reduced sensitivity from 83.7\% to 80.9\%, with a minor decrease in AUC (0.912 to 0.910). This suggests that clinical variables provide complementary cues to image-based discrimination, particularly for borderline cases.

Removing test-time augmentation (TTA) reduced AUC to 0.902 and sensitivity to 80.3\%, implying that TTA mainly contributes to robustness and recall. Notably, the fairness-gap metric decreased to 9.2\%, indicating that augmentation may be coupled with group disparities. Calibration or explicit fairness constraints may be needed to avoid improved overall performance at the cost of enlarged between-group differences. Removing Monte Carlo (MC) dropout slightly reduced sensitivity to 82.6\% and increased the fairness gap to 16.2\%, supporting the positive role of explicit uncertainty estimation in improving reliable outputs and cross-group consistency.

Overall, the ablation study not only quantifies each component’s contribution but also indicates that fairness and uncertainty mechanisms should be treated as integral to medical AI systems. Beyond AUC, models should be evaluated and optimized jointly for overall performance, missed-diagnosis risk, and between-group disparities.

\begin{table}[htbp]
    \centering
    \renewcommand{\arraystretch}{1.3} 
    \setlength{\tabcolsep}{7pt} 
    
    \caption{Ablation Study (Top 30\% Subset)}
    \label{tab:ablation_study}
    \begin{tabular}{lcccc}
        \hline
        \textbf{Method} & \textbf{AUC} & \textbf{Sens.} & \textbf{Spec.} & \textbf{Gap} \\
        \hline
        \textbf{Fair-Eye Net (Ours)} & \textbf{0.912} & \textbf{83.7\%} & \textbf{96.7\%} & \textbf{14.8\%} \\
        - Clinical Info & 0.910 & 80.9\% & 97.8\% & 11.0\% \\
        - TTA & 0.902 & 80.3\% & 98.9\% & 9.2\% \\
        - MC Dropout & 0.910 & 82.6\% & 96.7\% & 16.2\% \\
        \hline
    \end{tabular}
\end{table}

\subsection{Limitations and Future Work}
Although these results indicate utility for screening, grading, risk prediction, and dynamic warning, rigorous validation on larger multi-center external cohorts and in real clinical workflows remains necessary. Next, we will optimize monitoring/warning/intervention thresholds via cost-sensitive analysis and clinically grounded benefit metrics (e.g., referral burden and visual-function outcomes), and stress-test fairness calibration under cross-domain shifts (region, device, and population mix). We will also advance longitudinal modeling with time-to-event prediction and personalized follow-up interval recommendations to improve warning lead time and outpatient fit.

\section{Conclusion}
Fair-Eye Net integrates multimodal fusion, uncertainty-aware decision support, cross-domain robustness, and fairness constraints for glaucoma screening and longitudinal risk management. Experimental results suggest that the framework supports early detection, risk stratification, and monitoring while keeping high-stakes cases under clinician oversight. Future work will emphasize longer-term prospective validation and tighter integration with hospital information systems to enable sustainable, real-world deployment.

\bibliographystyle{IEEEtran}
\bibliography{refs}

\end{document}